\documentclass[authoryear]{elsarticle}
\makeatletter
\def\ps@pprintTitle{%
 \let\@oddhead\@empty
 \let\@evenhead\@empty
 \def\@oddfoot{}%
 \let\@evenfoot\@oddfoot}
\makeatother

\usepackage{amsmath}
\usepackage{amssymb}
\usepackage{amsfonts}
\usepackage{setspace}
\usepackage{rotating}
\usepackage{multirow}
\usepackage{graphicx}
\usepackage{color}
\usepackage{lineno}
\usepackage{framed}

\usepackage[colorlinks=true,allcolors=blue]{hyperref}

\usepackage[all]{xy}

\usepackage[hmargin=1in,vmargin=1in]{geometry}

\newcommand{\Real}{\mathbb R}

\newcommand{\x}{{\mathbf x}}
\newcommand{\w}{{\mathbf w}}

\newcommand{\vect}[1]{{\boldsymbol{\mathbf{#1}}}} 
\newcommand{\y}{{\vect y}}

\newcommand{\cut}[1]{} 

\DeclareMathOperator{\sign}{sign}

\makeatletter
\def\ps@pprintTitle{%
 \let\@oddhead\@empty
 \let\@evenhead\@empty
 \def\@oddfoot{}%
 \let\@evenfoot\@oddfoot}
\makeatother
\usepackage{hyperref}
\hypersetup{
    bookmarks=true,         
    unicode=false,          
    pdftoolbar=true,        
    pdfmenubar=true,        
    pdffitwindow=false,     
    pdfstartview={FitH},    
    pdftitle={RKS},         
    pdfauthor={Perez-Suay et al}, 
    pdfsubject={RKS},       
    pdfcreator={Perez-Suay et al},  
    pdfproducer={Perez-Suay et al}, 
    pdfkeywords={keyword1} {key2} {key3}, 
    pdfnewwindow=true,      
    colorlinks=true,        
    linkcolor=blue,         
    citecolor=blue,         
    filecolor=blue,         
    urlcolor=blue           
}

\journal{}

\begin{document}


\begin{frontmatter}
\title{Randomized Kernels for Large Scale\\
Earth Observation Applications}

\author[uv]{Adrian P\'erez-Suay\corref{cor1}}
\ead{adrian.perez@uv.es}
\author[uv]{Julia Amor\'os-L\'opez}
\author[uv]{Luis G\'omez-Chova}
\author[uv]{Valero Laparra}
\author[uv]{Jordi Mu{\~n}oz-Mar\'i}
\author[uv]{Gustau Camps-Valls}

\cortext[cor1]{Corresponding author}
\address[uv]{Image Processing Laboratory (IPL), Parc Cient\'ific, Universitat de Val\`encia, C/ Catedr\'atico Jos\'e Beltr\'an, 2. 46980 Paterna, Val\`encia, Spain. \url{adrian.perez@uv.es}, \url{http://isp.uv.es}}

\begin{abstract}
Current remote sensing applications of biophysical parameter estimation and image classification have to deal with an unprecedented big amount of heterogeneous and complex data sources. New satellite sensors involving a high number of improved time, space and wavelength resolutions give rise to challenging computational problems. Standard physical inversion techniques cannot cope efficiently with this new scenario. Dealing with land cover classification of the new image sources has also turned to be a complex problem requiring large amount of memory and processing time. In order to cope with these problems, statistical learning has greatly helped in the last years to develop statistical retrieval and classification models that can ingest large amounts of Earth observation data. 
Kernel methods constitute a family of powerful machine learning algorithms, which have found wide use in remote sensing and geosciences. However, kernel methods are still not widely adopted because of the high computational cost when dealing with large scale problems, such as the inversion of radiative transfer models or the classification of high spatial-spectral-temporal resolution data. This paper introduces an efficient kernel method for fast statistical retrieval of bio-geo-physical parameters and image classification problems. The method allows to approximate a kernel matrix with a set of projections on random bases sampled from the Fourier domain. 
The method is simple, computationally very efficient in both memory and processing costs, and easily parallelizable. 
We show that kernel regression and classification is now possible for datasets with millions of examples and high dimensionality. Examples on atmospheric parameter retrieval from hyperspectral infrared sounders like IASI/Metop; large scale emulation and inversion of the familiar PROSAIL radiative transfer model on Sentinel-2 data; and the identification of clouds over landmarks in time series of MSG/Seviri images show the efficiency and effectiveness of the proposed technique.
\end{abstract}


\end{frontmatter}

\section{Introduction}
\subsection{Remote sensing and the big data challenge}

Earth-observation (EO) satellites provide a unique source of information to address some of the challenges of the Earth system science~\citep{Berger12}. EO deals with the important objective of monitoring and modelling the processes on the Earth surface and their interaction with the atmosphere. To accomplish this ambitious goal, EO deploys both data acquired by remote sensing airborne and satellite sensors, as well as quantitative {\em in situ} measurements of biophysical parameters~\citep{CampsValls11mc}. Predictive models of biophysical parameters and classification of remotely sensed images are thus relevant outputs to monitor our Planet. 

In this context, current EO applications of image classification and biophysical parameter estimation, have to deal with an unprecedented big amount of heterogeneous and complex data sources. Spatio-temporally explicit quantitative methods are a requirement in a variety of Earth system data processing applications. Optical Earth observing satellites for example, endowed with a high temporal resolution, enable the retrieval and hence monitoring of climate and bio-geophysical variables~\citep{Dorigo2007,schaepman09}. The super-spectral Copernicus Sentinel-2 (S2)~\citep{Drusch2012} and the forthcoming Sentinel-3 mission~\citep{Donlon12}, as well as the planned EnMAP~\citep{Stuffler2007}, HyspIRI~\citep{Roberts2012}, PRISMA~\citep{Labate2009} and FLEX~\citep{Kraft2013}, will soon provide unprecedented data streams. Very high resolution (VHR) sensors like Quickbird, Worldview-2 and the recent Worldview-3~\citep{lpbc14} also pose big challenges to data processing. The challenge is not only attached to optical sensors. Infrared sounders, like the Infrared Atmospheric Sounding Interferometer (IASI)~\citep{TOURNIER2002} sensor on board the Metop satellite series, impose even larger constraints: the orbit time of Metop satellites (101 minutes), the large spectral resolution (8461 spectral channels between 645~cm$^{-1}$ and 2760~cm$^{-1}$), and the spatial resolution (60$\times$1530 samples) of the IASI instrument yield several hundreds of gigabytes of data to be processed daily. EO radar images also increased in resolution, and the current platforms, such as ERS-1/2, ENVISAT, RadarSAT-1/2, TerraSAR-X, and Cosmo-SkyMED give raise to extremely fine resolution data that call for advanced scalable processing methods. Besides, we should not forget the availability of the extremely large remote sensing data archives already collected by several past missions, such ENVISAT, Seviri/MSG, Cosmo-SkyMED, Landsat, or SPOT. 

These large-scale data problems require enhanced processing techniques that should be accurate, robust and fast. Standard physical inversion techniques and parametric classification algorithms cannot cope (nor adapt to) this new scenario efficiently. 
Over the last few decades a wide diversity of methods have been developed to tackle particular EO processing tasks, but only a few of them made it into operational processing chains, and many of them are only in its infancy. 

\subsection{Machine learning for Earth Observation data analysis}

In order to cope with these problems, statistical learning (also known as machine learning) has greatly helped in the last years to develop statistical retrieval and classification models that can ingest large amount of Earth observation data. Machine learning has become a standard paradigm for the analysis of remote sensing and geoscience data, at both local and global scales~\citep{CampsValls11mc}. Machine learning actually constitute a relevant alternative to parametric and physically-based models, which rely on established physical relations and implement complex combinations of scientific hypotheses, and 
give rise to too rigid solutions and eventual model discrepancies (see \citet{Berger12} and references therein). 

Alternatively, the framework of {\em statistical inference and machine learning} is concerned about developing {\em data-driven models} 
and they solely rely on the ``unreasonable effectiveness of data''~\citep{Halevy09}. The field has proven successful in many disciplines of Science and Engineering~\citep{Hastie09} and, in general, nonlinear and nonparametric model instantations typically lead to more flexible and improved performance over physically-based approximations. 

In the last decade, machine learning has attained outstanding results in the estimation of climate variables and related bio-geo-physical parameters at local and global scales, and on the classification of remote sensing images~\citep{CampsValls11mc}. Current operational vegetation products, like leaf area index (LAI), are typically produced with neural networks ~\citep{Bacour06, Duveiller2011,Baret2013}. Gross Primary Production (GPP) as the largest global CO$_2$ flux is estimated using ensembles of random forests and neural networks~\citep{Beer10,Jung11}. Similarly, the contribution of supervised classifiers has been improving the efficacy of the land cover/use mapping methods since the 1970s: Gaussian models such as Linear Discriminant Analysis ({LDA}) were replaced in the 1990s by non-parametric models able to fit the distribution observed in data of increasing dimensionality, which were later superseded by decision trees~\citep{Han96,Fri97} and then by neural networks (NN,~\citet{Bischof1992,bischof98,Bru99}).


\subsection{Kernel machines and random features for efficient EO data processing}

The last decade kernel methods emerged as a family of powerful machine learning algorithms, and found wide use in remote sensing and geosciences~\citep{CampsValls09wiley,CampsValls11mc}. In the last decade, a kernel method called support vector machines ({SVM},~\citet{huang02,campsieee04,Mel04b,foody04,campsvallstgars05}) was gradually introduced in the field, and quickly became a standard for image classification. Further SVM developments considered the simultaneous integration of spatial, spectral and temporal information~\citep{Benediktsson05,fauvel08,Pac08,Tuia09tgrs,cam08}, the richness of hyperspectral imagery~\citep{campsvallstgars05,Plaza09}, and exploiting the power of clusters of computers~\citep{Pla08b,Mun09b}. 
We observed a similar adoption of kernel machines for biophysical parameter retrieval: support vector regression showed high efficiency in modelling LAI, fCOVER and evapotranspiration~\citep{Yang06,Durbha07}, and kernel methods like Gaussian Processes (GPs)~\citep{Rasmussen06} recently provided excellent results in retrieving vegetation parameters~\citep{pasolli10,Verrelst12a,verrelst2013a,verrelst2013c,LazaroGredilla14,CampsValls16grsm}. 

However, kernel methods are still not widely adopted because of the high computational cost when dealing with large scale problems, such as the inversion of radiative transfer models or the classification of high spatial-spectral-temporal resolution data. Roughly speaking, given $n$ examples available to develop the models, kernel methods typically need to store in memory {\em kernel matrices} ${\bf K}$ of size $n\times n$ and to process them using standard linear algebra tools (matrix inversion, factorization, eigendecomposition, etc.). This is an important constraint that hamper its applicability to large scale EO data processing. 

In this paper, we introduce a kernel method for efficiently approximate kernels, that make nonlinear classification possibly with millions of examples. We will focus on the two most relevant EO data problems: statistical retrieval of bio-geo-physical parameters and image classification problems. The method allows to approximate a kernel matrix ${\bf K}$ with a set of random bases sampled from the Fourier domain. The method is simple, computationally very efficient in both memory and processing costs, and easily parallelizable through standard divide-and-conquer strategies as the ones proposed in~\citep{Wainwright13}. 

The contributions of this paper are: (1) the introduction to the remote sensing community of this new efficient method to perform nonlinear regression and classification; (2) the extension of the method to work with other than Fourier bases, such as wavelets, stumps, and Walsh expansions that can cope with other data characteristics; and (3) to give experimental evidences in several illustrative and challenging problems in EO data processing: parameter retrieval, model inversion, and remote sensing image classification. In particular, we will show that kernel regression/classification is now possible for datasets with millions of examples and high dimensionality. The efficiency and effectiveness of the technique is illustrated for atmospheric parameter retrieval from hyperspectral infrared sounders like IASI, large scale emulation and inversion of the familiar PROSAIL radiative transfer model on Sentinel-2 data, and the identification of clouds over landmarks in time series of MSG/Seviri images. In addition, we will show that the method is very simple to implement, computationally very efficient in both memory and processing cost, and easily parallelizable for operational services and product generation. 

\subsection{Outline}

The remainder of the paper is organized as follows. Next Section 2 briefly reviews the field of kernel methods and the proposed RKS approximation, discusses implementation issues, gives intuition about the involved parameters in the model, and introduces the different novel extensions introduced in this work. Section 3 presents and discusses the experimental results in three challenging problems of bio-geo-physical parameter retrieval and data classification. 
Section 4 concludes the paper with some discussion and remarks, outlines the future work, and notes some research opportunities in related EO fields.

\section{Random feature kernels for large scale EO data processing}

Kernel methods constitute an appropriate framework to approach many statistical inference problems~\citep{ShaweTaylor04}. In the last decade these methods have replaced other techniques in many fields of science and engineering, and have become the new standard in remote sensing data analysis~\citep{CampsValls09,CampsValls11mc}. Kernel methods allow treating in the very same framework different problems, from feature extraction~\citep{Arenas13spm} to classification~\citep{Camps-Valls14} and regression~\citep{CampsValls10eumtgrs}. The fundamental building block of the theory of kernel learning is the {\em kernel function}, which compares multidimensional data objects. In a nutshell, given $n$ data points, all kernel methods have to operate with a squared (eventually huge) matrix of size $n\times n$, which contains all pairwise sample similarities. Designing an appropriate kernel function that captures data dependencies is, nevertheless, not easy in general. Many approaches have been followed so far to tackle this problem: from learning the metric implicit in the kernel~\citep{weinberger07metricregression,weinberger08fast} 
to learning compositions of simpler kernels~\citep{Rakotomamonjy08,DuvLloGroetal13}.
Selecting and optimizing a kernel function is very challenging even with moderate amounts of data. Many efforts have been done to deliver large-scale versions of kernel machines able to work with several thousands of examples~\citep{Bottou02}. 
They typically resort to reduce the dimensionality of the problem by decomposing the kernel matrix using a subset of bases: for instance using Nystr\"om eigendecompositions~\citep{Kumar12}, sparse and low rank approximations~\citep{fine01efficient,Arenas13spm}, or smart sample selection~\citep{bordes2005fast}. However, there is no clear evidence that these approaches work in general, given that they are simple heuristic approximations to the kernel. 

\subsection{From linear to kernel least squares regression and classification}

Here we start by reviewing the standard linear regression (LR)~\citep{Geladi86} and its kernel version, the kernel ridge regression (KRR)~\citep{scholkopf02,ShaweTaylor04}. Note that KRR is also known as Least Squares SVM (LS-SVM), and it shares formulation with Gaussian processes (GPs)~\citep{Rasmussen06,CampsValls16grsm}. The main difference between KRR and GPs is the parameter training procedure: in KRR the training is usually done by cross-validation and in GPs the hyperparameters are inferred using gradient descent procedures on the marginal maximum likelihood. Reviewing LR and KRR formulation will help to understand the ideas behind the proposed RKS. 


Let ${\bf x}_i\in {\bf \mathbb{R}}^d$ (inputs) and ${\bf y}_i\in {\bf \mathbb{R}}^o$ (outputs), where $i=1,...,n$ indicates the index of the $n$ training samples. In the LR case we want to fit a linear function to predict the output sample ${\bf y}_\ast$ given an input sample ${\bf x}_\ast\in\Real^d$, i.e. ${\bf y}_\ast = {\bf W}^\top{\bf x}_\ast$ (we assume here that both input and output data are bias corrected). In order to fit the weights ${\bf W}$ we can use the training samples (where we have access to ${\bf y}_i$ examples) and perform the least squares solution: ${\bf W} = ({\bf X}^\top{\bf X})^{-1}{\bf X}^\top{\bf Y}$, where ${\bf X}$ and ${\bf Y}$ contain all the training input and output samples in a matrix form respectively, i.e. {\bf Y} is $n \times o$ and {\bf X} is $n \times d$, thus {\bf W} is $d \times o$. Note that in general the inversion of the matrix ${\bf X}^\top{\bf X}$ could be unstable, which is typically solved via Tikhonov's regularization, and then ${\bf W} = ({\bf X}^\top{\bf X} + \lambda {\bf I})^{-1}{\bf X}^\top{\bf Y}$. Either case, linear regression is computationally very efficient, as it only implies inverting a $d\times d$ matrix. Note that under a Bayesian interpretation, parameter $\lambda$ is related to the amount of noise in the input data. In our case, we fit the parameter $\lambda$ by minimizing the prediction error in a out-of-sample test set via a standard cross-validation procedure. 

In the KRR case, we want to perform a linear least squares regression in a Hilbert space, ${\mathcal H}$, of very high (possibly infinite) dimensionality $D_{\mathcal H}$, where samples have been mapped through a mapping $ \boldsymbol{\phi}(\cdot)$. In matrix notation, the model is given by ${\bf Y} = \boldsymbol{\Phi}{\bf W}_{\mathcal H}$, where now the weights are defined in the representation space of unknown dimensionality, ${\bf W}_{\mathcal H}\in{\mathcal H}$. Then, as in the regularized linear regression setting, we want to minimize the regularized squared loss function in ${\mathcal H}$,
$${\mathcal L}_p = \|{\bf Y} - \boldsymbol{\Phi}{\bf W}_{\mathcal H}\|^2 + \lambda \|{\bf W}_{\mathcal H}\|^2,$$ 
with respect to model weights ${\bf W}_{\mathcal H}$ and the regularization term $\lambda$. To fit $\lambda$ we are going to follow the same procedure as in LR, a classical cross-validation with the training samples. To find the solution for ${\bf W}_{\mathcal H}$ first we take derivatives with respect to ${\bf W}_{\mathcal H}$ and equating them to zero, by doing so we obtain ${\bf W}_{\mathcal H} = (\boldsymbol{\Phi}^\top\boldsymbol{\Phi} + \lambda {\bf I})^{-1} \boldsymbol{\Phi}^\top {\bf Y}$, where $\boldsymbol{\Phi}$ is the matrix of mapped samples, $[\boldsymbol{\phi}({\bf x}_1)^\top,\boldsymbol{\phi}({\bf x}_2)^\top,...,\boldsymbol{\phi}({\bf x}_n)^\top]$, whose size is $n\times D_{\mathcal H}$. Note that this problem is not solvable as the inverse runs on matrix $\boldsymbol{\Phi}^\top\boldsymbol{\Phi}$ which is of size $D_{\mathcal H}\times D_{\mathcal H}$, and $\boldsymbol{\Phi}$ is in principle unknown. However, by applying the Representer's theorem we can express the solution as a linear combination of mapped samples, ${\bf W}_{\mathcal H}$ = $\boldsymbol{\Phi}^\top\boldsymbol{\Lambda}$, and then the solution is expressed as a function of the dual weights $\boldsymbol{\Lambda}\in\Real^{n\times o}$ (one per sample and variable), $\boldsymbol{\Lambda} = (\boldsymbol{\Phi}\boldsymbol{\Phi}^\top + \lambda {\bf I})^{-1} {\bf Y}$. Now the problem is solvable as we only need to compute the inverse of the (regularized) Gram matrix ${\bf K} := \boldsymbol{\Phi}\boldsymbol{\Phi}^\top$ of size $n\times n$. Even though the mapping is unknown, one can replace this inner product matrix with a similarity matrix between samples, which is known as the {\em kernel matrix}, ${\bf K}$. 

We finally need to show that we never actually require access to the mapped feature vectors into ${\mathcal H}$. In practice, for a new test example ${\bf x}_\ast$, we only want the predicted value (${\bf y}_\ast$), which is computed by projecting it onto the solution ${\bf W}_{\mathcal H}$, and then  replacing the dot product with the kernel function:
\begin{eqnarray}
\hat {\bf y}_\ast = \boldsymbol{\phi}_\ast {\bf W}_{\mathcal H} = \boldsymbol{\phi}_\ast\boldsymbol{\Phi}^\top\boldsymbol{\Lambda} = {\bf K}_\ast\boldsymbol{\Lambda},
\end{eqnarray}
where the matrix ${\bf K}_\ast$ contains the similarities between the test example $\x_\ast$ and all training samples, ${\bf X}$. The important message here is of course that we only need access to the kernel function $K(\cdot,\cdot)$ that measures the similarity between two feature vectors, not the nonlinear mapping function $\boldsymbol{\phi}(\cdot)$. Examples of typical kernel functions are the linear $K$(${\bf x}_i$,${\bf x}_j$) = ${\bf x}_i^\top {\bf x}_j$, the polynomial $K$(${\bf x}_i$,${\bf x}_j$) = ${(a{\bf x}_i^\top {\bf x}_j+b)^p}$, and the one used here, the Gaussian Function (Radial Basis Function, RBF) kernel $K({\bf x}_i,{\bf x}_j) = \exp(-\| {\bf x}_i - {\bf x}_j\|^2/(2\sigma^2)$). Therefore, we only have two free parameters to tune: the regularization parameter $\lambda$ and the kernel parameter $\sigma$. As for the $\lambda$ parameter, we used a cross-validation strategy for optimizing $\sigma$. For the interested reader, a MATLAB implementation of KRR and other regression algorithms can be found at our web page \url{http://isp.uv.es/}. 

The KRR can be readily used for binary classification as well, in which case is known as the least squares SVM. As noted in \citep{Suykens1999}, the least squares classification problem is essentially the same as the regression problem by (1) considering the signal model $f(\x_i) = \text{sign}[\w^\top\boldsymbol{\phi}(\x_i)+b]$; and (2) introducing equality constraints $\y_i(\w^\top\boldsymbol{\phi}(\x_i)+b)=1-{\bf e}_i$, where ${\bf e}_i$ is the error variable, $i=1,\ldots,n$. The optimization problem is exactly the same as for regression, and one only needs to encode the binary class labels as $\{-1,+1\}$ and then apply the $\text{sign}[\cdot]$ operator on the model predictions.

In conclusion, the KRR/LS-SVM can be seen as a regularized linear regression in a (possibly) infinite feature space. For doing this regression, one only needs to compute the kernel (Gram) matrix ${\bf K}\in\Real^{n\times n}$, and solve the normal equations. This can be a huge computational challenge depending on the amount of available training data examples $n$. This is why very often kernel methods in remote sensing did not make it to be operational in problems with more than a few thousand labeled training data. In addition, note that in KRR all the training examples receive a weight, and in the test (prediction) phase, one needs to compare (compute the similarity through the kernel function) of all test data to all training data, ${\bf K}_\ast$. This is also a problem when millions of instances arrive in the production phase. Both problems can be addressed by explicitly defining the feature mapping, as we will see next in the proposed RKS method.

\subsection{Approximating kernels with projections on random features} 

Instead that adopting a standard kernel function, in this paper we explore an alternative pathway: rather than {\em optimization} we will follow {\em randomization}. While odd at a first glance, the approach has surprisingly yielded competitive results in last years, being able to exploit many samples at a fraction of the computational cost. Besides its practical convenience, the approximation of the kernel with random bases is also theoretically consistent.  The seminal work in~\citep{Rahimi07} presented the randomization framework. Given a sample set $\{\x_i\in\Real^d|i=1,\ldots,n\}$, the idea is to approximate the kernel function with an empirical kernel mapping of the form:
$$K(\x_i,\x_j) = \boldsymbol{\phi}(\x_i)^\top\boldsymbol{\phi}(\x_j) \approx {\bf z}(\x_i)^\top {\bf z}(\x_j),$$
where the {\em implicit} mapping $\boldsymbol{\phi}(\cdot)$ is replaced with an {\em explicit} (low-dimensional) feature mapping ${\bf z}(\cdot)$ of dimension $D$. 
Consequently, one can simply transform the input with ${\bf z}$, and then apply fast linear learning methods to approximate the corresponding nonlinear kernel machine. This approach not only provides extremely fast learning algorithms, but also good performance in the test phase. 
The question now is how to construct efficient and sensible ${\bf z}$ mappings. The work in~\citep{Rahimi07} also introduced a particular technique to do so. 

\begin{figure}[h!]
\begin{center}
\begin{tabular}{cc}
RBF, ideal & RKS, $D=1$  \\
\includegraphics[width=6cm]{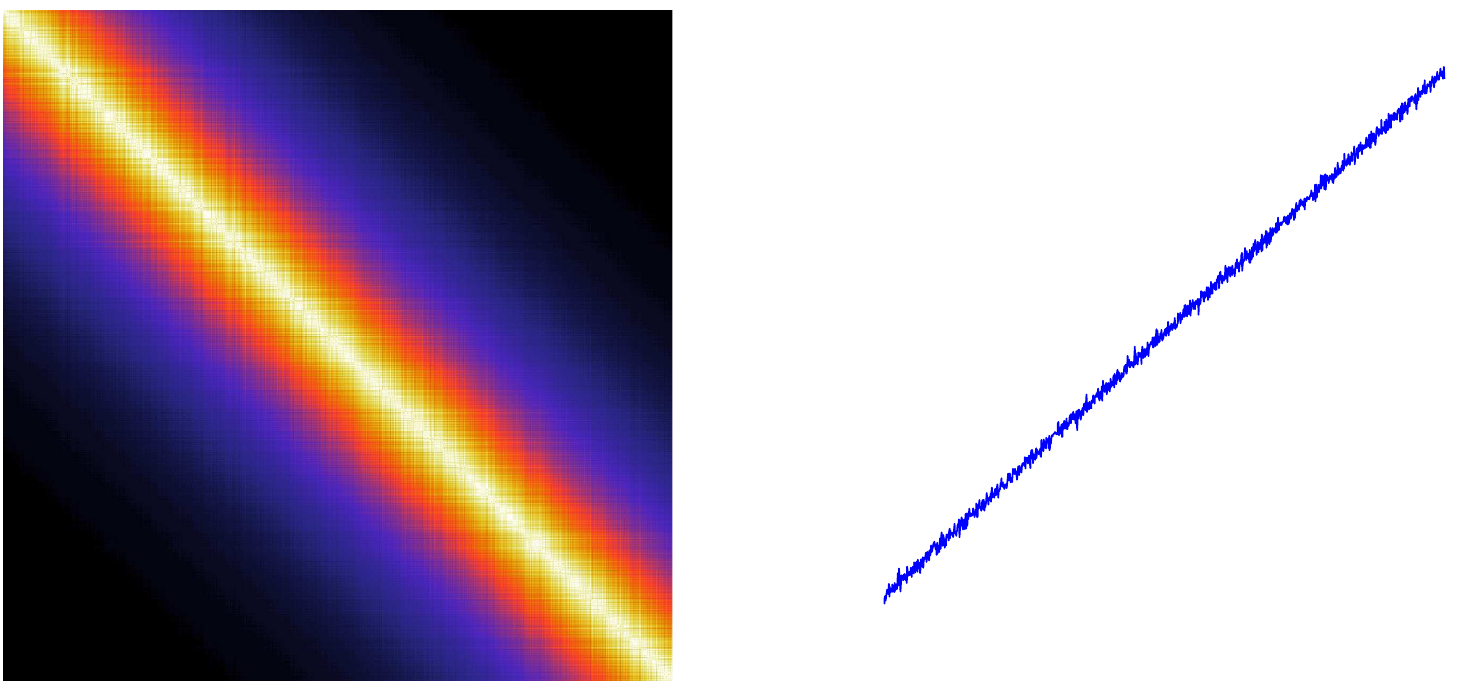} & \includegraphics[width=6cm]{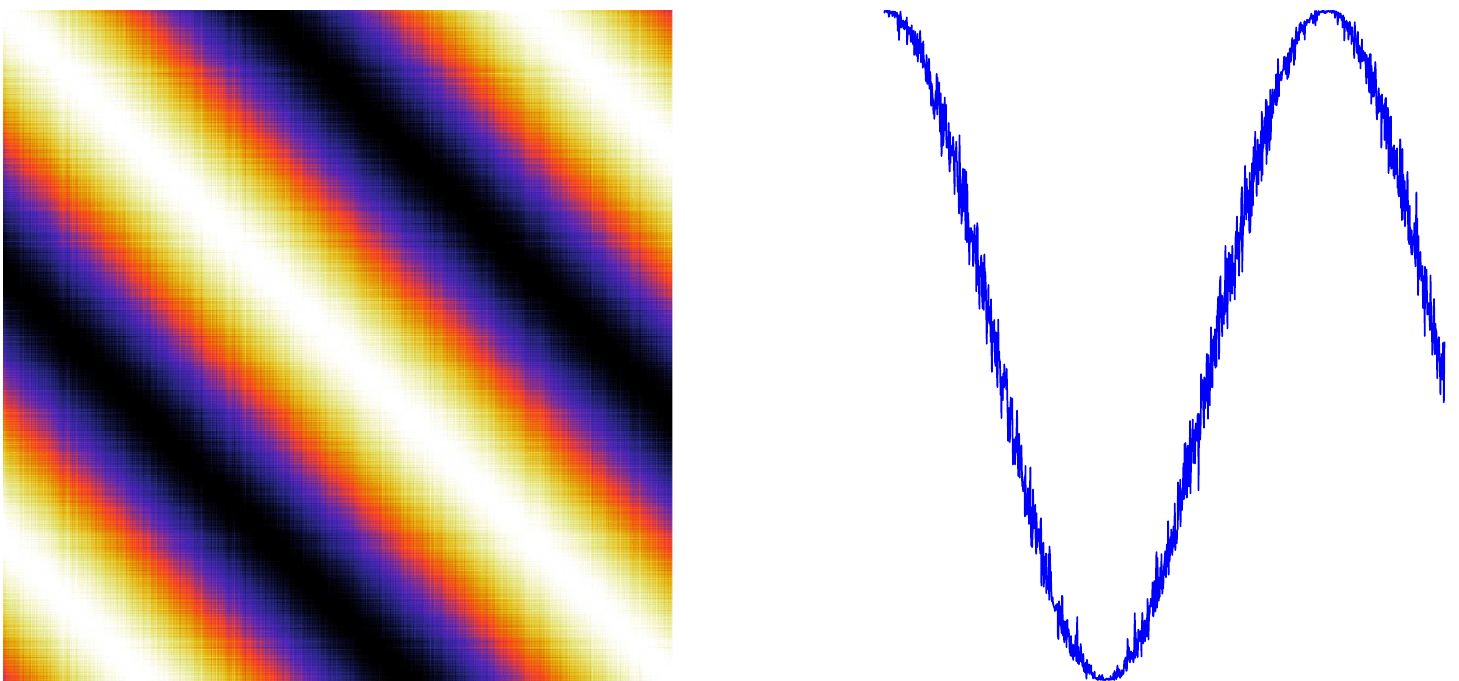} \\
RKS, $D=5$ & RKS, $D=1000$  \\
\includegraphics[width=6cm]{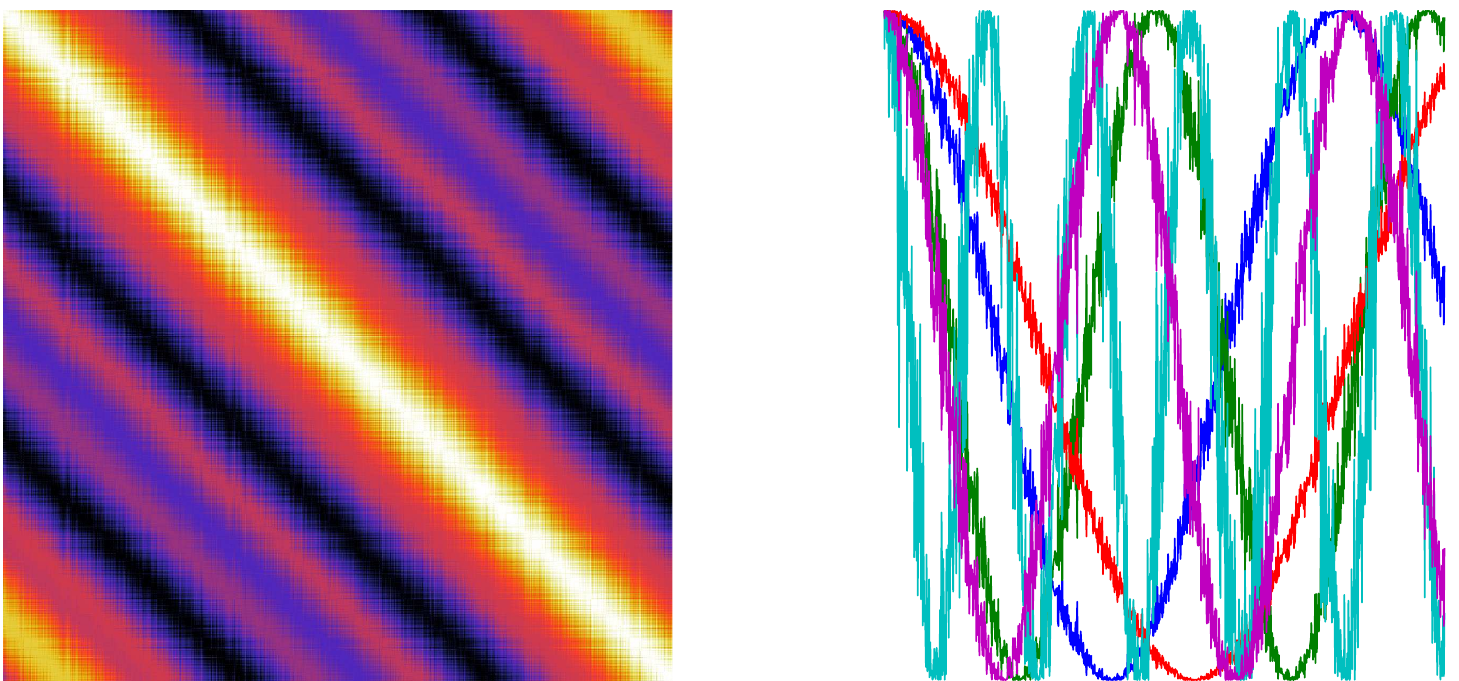}      & \includegraphics[width=6cm]{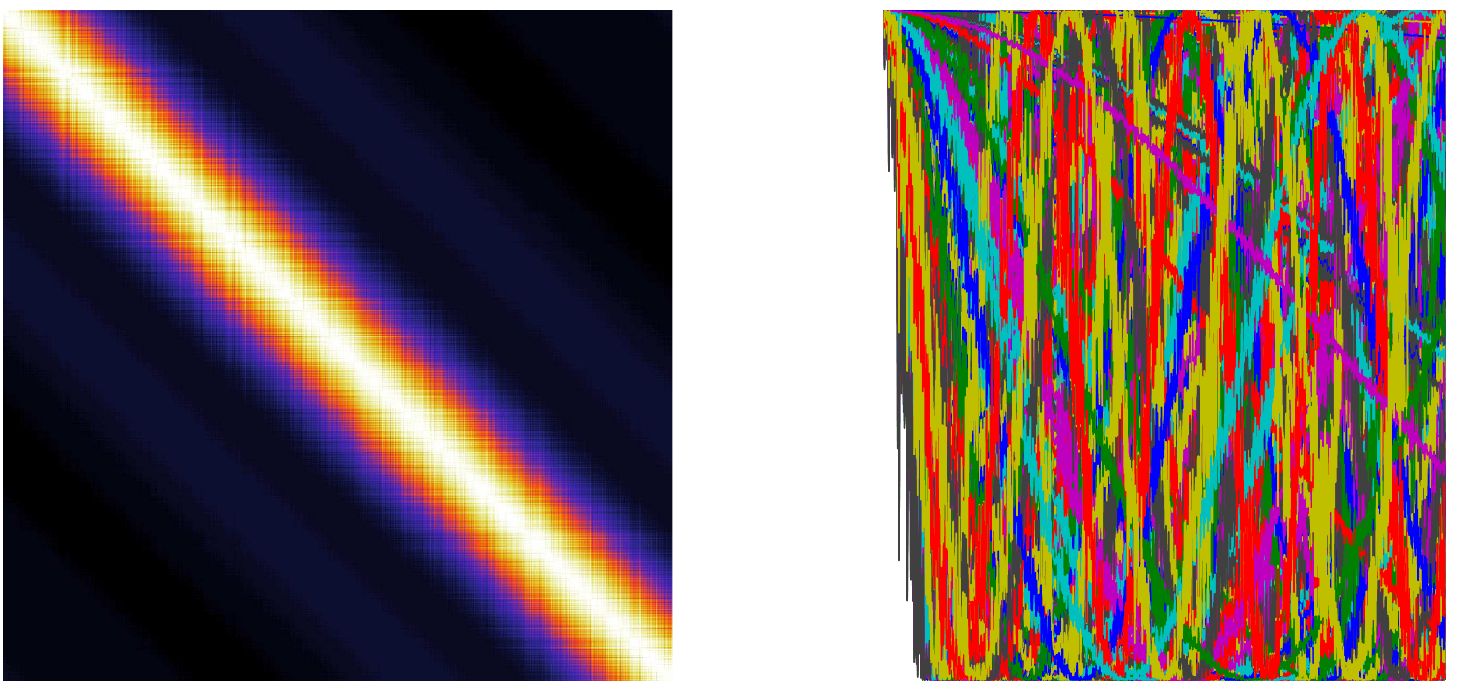}
\end{tabular}
\end{center}
\caption{Illustration of the effect of randomly sampling $D$ bases from the Fourier domain on the kernel matrix. With sufficiently large $D$, the kernel matrix generated by RKS approximates that of the RBF kernel, at a fraction of the time. }
\label{fig:example}
\end{figure}

The method exploits a classical definition in harmonic analysis~\citep{Rahimi07}, by which a continuous kernel $K(\x,\y)=K(\x-\y)$ on $\Real^d$ is positive definite if and only if $K$ is the Fourier transform of a non-negative measure. If a shift-invariant kernel $K$ is properly scaled, its Fourier transform $p(\boldsymbol{\omega})$ is a proper probability distribution. Defining the function $C_\omega(\x) = e^{j\boldsymbol{\omega}^\top \x}$, we obtain
$$K(\x-\y) = \int_{\Real^d}p(\boldsymbol{\omega})e^{j\boldsymbol{\omega}^\top (\x-\y)}d\boldsymbol{\omega} = {\mathbb E}_\omega[C_\omega(\x)C_\omega(\y)^*],$$
so $C_\omega(\x)C_\omega(\y)^*$ is an unbiased estimate of $K(\x-\y)$ when $\boldsymbol{\omega}$ is drawn from $p$. 
In our case, both $p(\boldsymbol{\omega})$ and $K(\x-\y)$ are real valued, what allows us to substitute the complex exponentials by cosines and to use $z_\omega(\x)^\top z_\omega(\y)$, where $z_\omega(\x) = \sqrt{2}cos(\boldsymbol{\omega}^\top\x+b)$, as an estimator of $K(\x-\y)$ as long as $\boldsymbol{\omega}$ is drawn from $p(\boldsymbol{\omega})$ and $b$ is drawn uniformly from $[0,2\pi]$.
Also note that $z_\omega(\x)^\top z_\omega(\y)$ has expected value $K(\x,\y)$ because of the sum of angles formula. Now, one can lower the variance of the estimate of the kernel by concatenating $D$ randomly chosen ${\bf z}_\omega$ into one $D$-dimensional vector ${\bf z}$ and normalizing each component by $\sqrt{D}$. An illustrative example of how to approximate the kernel $K$ with random bases is given in Fig.~\ref{fig:example}.

\subsection{RKS in practice for classification and regression problems} 

The RKS algorithm reduces to two simple steps: first draw $D$ i.i.d. samples $\boldsymbol{\omega}_1,\ldots,\boldsymbol{\omega}_D\in \Real^d$ from $p$, and $b_1,\ldots,b_D\in\Real$ from the uniform distribution $[0,2\pi]$; and then construct the low-dimensional feature map ${\bf z} = \sqrt{\frac{2}{D}}~[\cos(\boldsymbol{\omega}_1^\top \x+b_1),\ldots,\cos(\boldsymbol{\omega}_D^\top \x+b_D)]$. One could actually use the explicit mapping ${\bf z}$ to approximate the kernel function $K(\x_i,\x_j)\approx{\bf z}_i^\top {\bf z}_j$ and its associated kernel matrix, ${\bf K}={\bf Z}{\bf Z}^\top$, where ${\bf Z}:=[{\bf z}_1,\ldots,{\bf z}_n]^\top\in\Real^{n\times D}$ for optimization and prediction. However, this would result in the same computational burden. The RKS proceeds in solving the primal problem directly: one first replaces the implicitly mapped data $\boldsymbol{\Phi}$ with the explicit maps ${\bf Z}$, and solves the normal equations as for LR. This procedure returns the following RKS solution:
$${\bf W} \approx ({\bf Z}^\top{\bf Z})^{-1}{\bf Z}^\top{\bf Y},$$
which is computationally equivalent to solving the least squares linear regression model. The cost of the RKS reduces to invert matrix ${\bf Z}^\top{\bf Z}$ of size $D\times D$, which can be very convenient  for large scale problems compared to the standard KRR/LS-SVM procedure that involves matrices of size $n\times n$. This is particularly important as in many current problems in EO data processing, $n$ is large but feature redundancy is typically present (either spatial, spectral, temporal) which justifies the use of relatively low number of random features $D$ to project data onto, and thus reduces the memory requirements drastically. The latter is in turn important for test phases involving large data streams. Efficiency in both speed and memory requirements, and for both training and testing phases, are summarized in Table~\ref{tab:cost}. 

\begin{table}[h!]
\renewcommand{\tabcolsep}{2pt}
\caption{Computational time and memory costs for different linear, (approximate) kernel methods and random feature kernels in problems with $d$ dimensions, $D$ features, and $n$ samples.\label{tab:cost}}
\begin{center}
\begin{tabular}{|l|llll|}
\hline
\hline
Method & Train time & Test time & Train mem & Test mem \\
\hline
LR~\citep{Geladi86}           & ${\mathcal O}(d^2 n)$   &  ${\mathcal O}(d^2)$  &  ${\mathcal O}(d^2)$  &  ${\mathcal O}(d)$ \\
\hline
Naive~\citep{ShaweTaylor04}          & ${\mathcal O}(n^2 d)$   &  ${\mathcal O}(n d)$  &  ${\mathcal O}(n d)$  &  ${\mathcal O}(n d)$\\
\hline
Low Rank~\citep{fine01efficient}       & ${\mathcal O}(n D d)$   &  ${\mathcal O}(D d)$  &  ${\mathcal O}(D d)$  &  ${\mathcal O}(D d)$ \\
\hline
RKS~\citep{Rahimi07}            & ${\mathcal O}(D d n)$   &  ${\mathcal O}(D d)$  &  ${\mathcal O}(D d)$  &  ${\mathcal O}(D d)$ \\
\hline
\hline
\end{tabular}
\end{center}
\end{table}


Finally, we would like to highlight that the RKS algorithm can actually exploit other approximating functions besides Fourier expansions. Note that actually any shift-invariant kernel, i.e. $K(\x,\y) = K(\x-\y)$, can be represented using random cosine features. Randomly sampling distribution functions impacts the definition of the corresponding reproducing kernel Hilbert space (rkHs): sampling the Fourier bases with $z_\omega(\x) =\sqrt{2}\cos(\boldsymbol{\omega}_o^\top \x+b)$ actually leads to the Gaussian RBF kernel $K(\x,\y)=\exp(-\|\x-\y\|^2/(2\sigma^2))$, while a random stump (i.e. sigmoid-shaped functions) sampling defined by $z_\omega(\x) =\sign(\x-\boldsymbol{\omega})$ leads to the kernel $K(\x,\y)=1-\frac{1}{a}\|\x-\y\|_1$. Another possibility is to resort to binning bases functions, which partition the input space using an axis-aligned grid, and assign a binary indicator to each partition, which is shown to approximate a Laplacian kernel, $K(\x,\y)=\exp(-\|\x-\y\|_1/(2\sigma^2))$~\citep{Rahimi07}. In this paper, we will also explore the possibility of Walsh and the Gabor basis functions widely used in signal and image processing. 

\section{Experimental results}
This section presents experimental results on the use in RKS in several remote sensing applications: atmospheric parameter retrieval from IASI infrared sounding data; emulation and inversion of the PROSAIL radiative transfer model; and cloud detection over landmarks in MSG/SEVIRI image time series. 
\subsection{Experiment 1: Atmospheric parameter retrieval from MetOp/IASI infrared sounding data}
In this first experiment, we exploit random feature kernels in a challenging regression problem in remote sensing: the estimation of atmospheric profiles from large scale hyperspectral infrared sounders. 
Temperature and water vapor are atmospheric parameters of high importance for weather forecast and atmospheric chemistry studies \citep{Liou2002,Hilton2009}. 
Observations from spaceborne high spectral resolution infrared sounding instruments can be used to calculate the profiles of such atmospheric parameters with unprecedented accuracy and vertical resolution \citep{Huang1992}. In this work, we focus on  the Infrared Atmospheric Sounding Interferometer (IASI) onboard Metop. The use of Metop data in Numerical Weather prediction (NWP) accounts for 40\% of the impact of all space based observations in NWP forecasts. 


Products obtained from IASI data are a significant improvement in the quality of the measurements used for meteorological models. In particular, IASI collects rich spectral information to derive temperature and moisture profiles, which are essential to the understanding of weather and to derive atmospheric forecasts. The sensor provides infrared spectra with high resolution between 645~cm$^{-1}$ and 2760~cm$^{-1}$, from which temperature and humidity (or related dew point temperature) profiles with high vertical resolution and accuracy are derived. Additionally, it is used for the determination of trace gases such as ozone, nitrous oxide, carbon dioxide and methane, as well as land and sea surface temperature and emissivity and cloud properties. In summary, IASI provides radiances in $8461$ spectral channels, between $3.62$ and $15.5$~$\mu$m with a spectral resolution of $0.5$~cm$^{-1}$ after apodization \citep{Simeoni1997,Chalon01}. Its spatial resolution is $25$~km at nadir with an Instantaneous Field of View (IFOV) size of $12$~km at an altitude of $819$~km. This huge data dimensionality typically requires simple and computationally efficient processing techniques that can exploit the wealth of available observations provided by ECMWF re-analysis. 

Actually, EUMETSAT, NOAA, NASA and other operational agencies are continuously developing product processing facilities to obtain L2 atmospheric profile products from infrared hyperspectral radiance instruments, such as IASI. One of the retrieval techniques commonly used in L2 processing is based on the canonical linear regression (LR), which is a valuable and very computationally efficient method. It consists of performing a canonical least squares linear regression on top of the data projected onto the first principal components or Empirical Orthogonal Functions (EOF) of the measured brightness temperature spectra (or radiances) and the atmospheric state parameters. To further improve the results of this scheme for retrieval, {\em nonlinear statistical retrieval methods} can be applied as an efficient alternative to more costly optimal estimation (OE) schemes. These methods have proven to be valid in retrieval of temperature, dew point temperature, and ozone atmospheric profiles when the original data are used \citep{CampsValls10eumtgrs}.  


In this experiment, we followed the same procedure as in \citep{CampsValls10eumtgrs} where LR and KRR were applied using hyperpixels (i.e. all the spectral components at a particular spatial position) of IASI data to predict temperature and dew point temperature at different pressure levels. First, the dimensionality of the data is reduced to $100$ principal components by using the classical principal component analysis (EOF/PCA) transformation in the spectral domain. Then, LR and KRR models are trained using a fraction of the data and the rest of the data is employed to assess the models' performance. In addition to the standard LR and KRR, we also incorporate the proposed RKS method for retrieval, which allows us to train nonlinear regression efficiently.

\begin{figure}[t!]
\renewcommand{\tabcolsep}{1pt}
\begin{center}
\begin{tabular}{ccc}
(a) & (b) & (c) \\
\includegraphics[width=5.5cm]{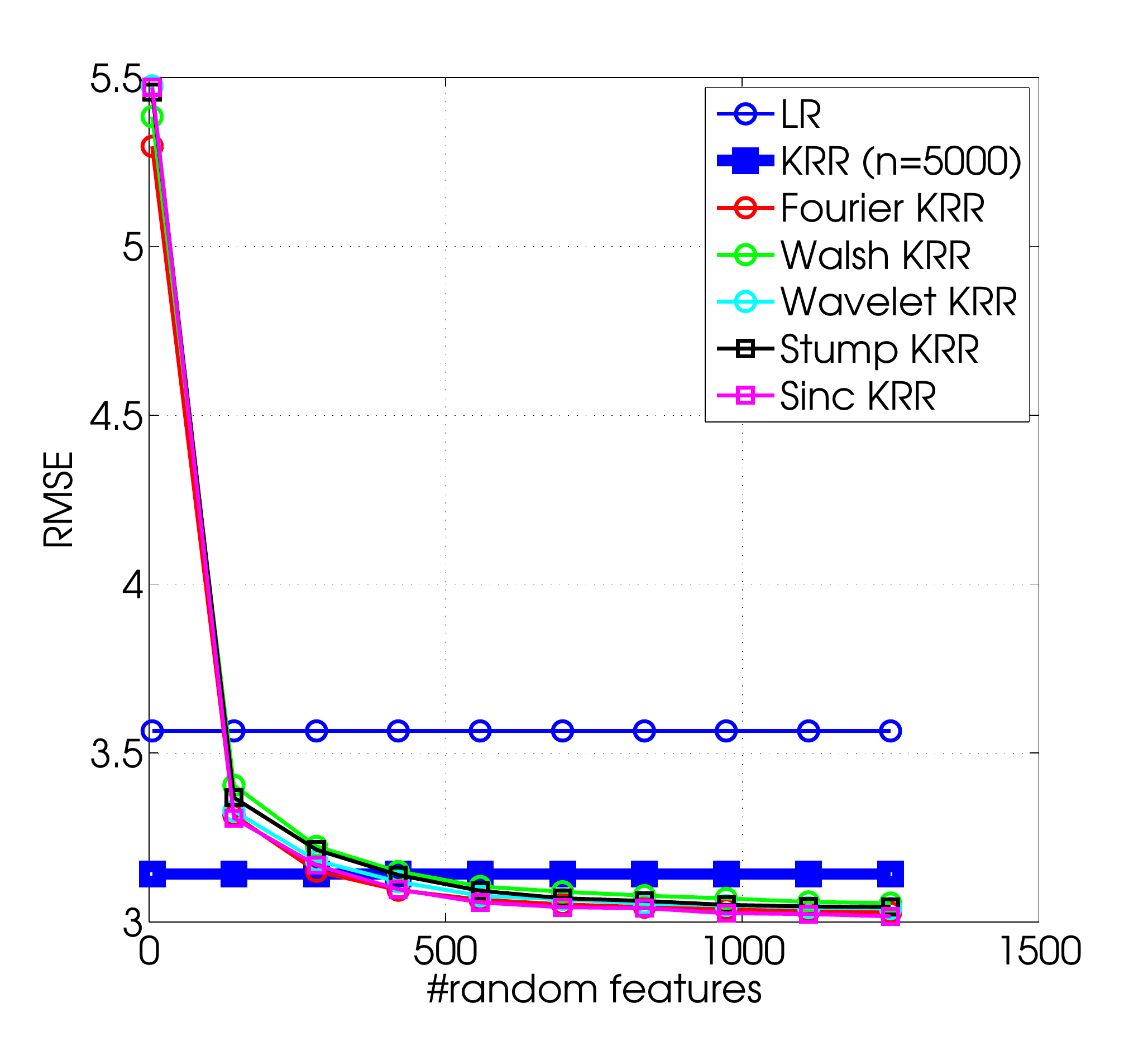} &
\includegraphics[width=5.5cm]{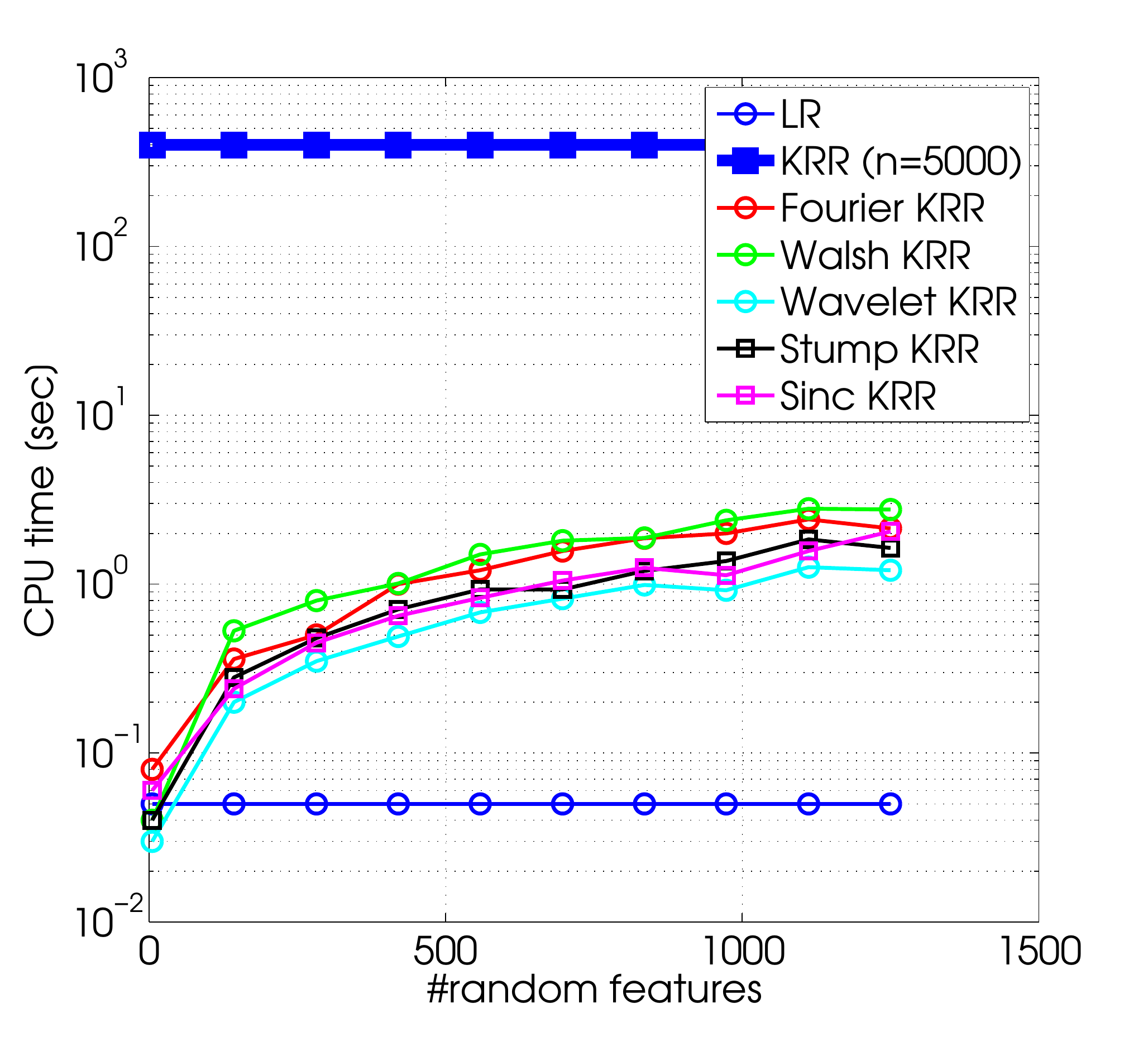}  &
\includegraphics[width=5.5cm]{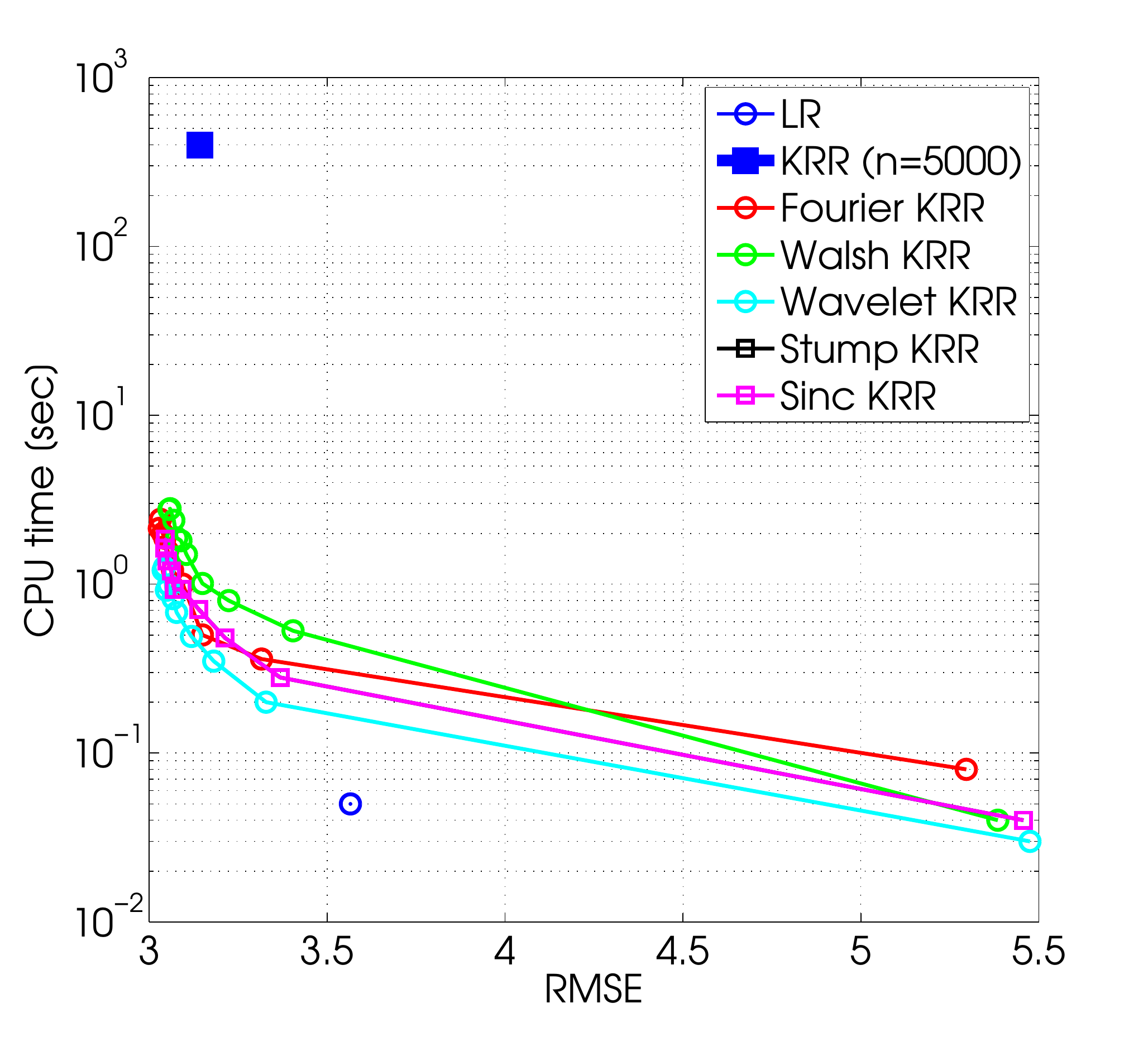}
\end{tabular}
\end{center}
\caption{Results of the RKS approach for different random sinks: (a) RMSE [K] and (b) training time [sec] versus the number of random features drawn; and (c) RMSE [K] versus training time [sec].}
\label{fig:rks}
\end{figure}

Figure~\ref{fig:rks} shows results for the prediction of the temperature atmospheric profile. We trained LR and KRR models using 5000 examples from an IASI orbit (2008-07-17), while the RKS approximations were trained with an ensemble of 100,000 examples. All models were then tested on the same independent test set of 20,000 examples. Experiments were performed in a standard laptop using MATLAB on an Intel 3.3 GHz processor with 8 GB RAM memory under Ubuntu 14.4. Figure~\ref{fig:rks}(a) shows that LR cannot cope with the nonlinearity of the problem, which can be addressed by using the kernel least squares regression method, KRR. However, training the KRR with more than 5000 samples turns out to be hard in regular machines. Using RKS instead is beneficial. It is actually observed that a sufficiently large number of randomly sampled bases for kernel approximation can improve the results in terms of accuracy and computational efficiency: in this case $>600$ random features were enough to beat the full 5000-samples KRR. The big leap in computational cost is observed in Fig.~\ref{fig:rks}(b) (note the log-scale). A trade-off comparison in Fig.~\ref{fig:rks}(c) reveals that the best accuracy-cost compromise in this particular example is to sample from the traditional squared-shaped Haar wavelet.

\subsection{Experiment 2: Emulation and inversion of the PROSAIL radiative transfer model for Sentinel-2}

The second experiment deals with both the {\em statistical emulation and inversion} of the familiar PROSAIL radiative transfer model. 
PROSAIL is the combination of the PROSPECT leaf optical properties model and the SAIL canopy bidirectional reflectance model. PROSAIL has been used to develop new methods for retrieval of vegetation biophysical properties. Essentially, PROSAIL links the spectral variation of canopy reflectance, which is mainly related to leaf biochemical contents, with its directional variation, which is primarily related to canopy architecture and soil/vegetation contrast. This link is the key to the simultaneous estimation of canopy biophysical/structural variables for applications in agriculture, plant physiology, and ecology at different scales. PROSAIL has become one of the most popular radiative transfer tools due to its ease of use, robustness, and consistent validation by lab/field/space experiments over the years. Our aim in this set of experiments is to both mimic and invert PROSAIL with kernel methods: these two processes imply solving the forward and inverse problems involving multidimensional large-scale datasets.


A standard approach in vegetation parameter retrieval consists of inverting PROSAIL. This {\em hybrid} approach essentially implies simulating radiances using PROSAIL for a set of state vectors and observation conditions. The inversion can be done in several ways, either using numerical optimization, look-up-tables, or statistical approaches (as we are interested herein). Either way, the critical question about the representativity of the created dataset still remains: typically no more than a few thousand points are generated using PROSAIL with the hope that these variable-radiances pairs explain the problem variability well. Even for fast RTMs such as PROSAIL, generating millions of pairs is challenging computationally. In recent years, machine learning techniques have been used not only for model inversion but for {\em emulation} of RTMs, i.e. statistical models act as fast approximations to complex physical models. This approach has a long and successful story in statistics~\citep{OHagan78,sacks1989design,Kennedy2001}, but efficient implementations were not accessible to the large audience because of the high computational burden involved in training the algorithms. Emulators are essentially surrogate models or metamodels: they are generally orders of magnitude faster than the original RTM, and can then be used \emph{in lieu} of it, opening the door to more advanced biophysical parameter estimation methods, using e.g. data assimilation (DA) concepts~\citep{quaife2008assimilating,lewis2012earth}.

Importantly for our interests, we should note that, once trained, the emulator can be used to generate new reflectances from new state vectors extremely fast: note that the cost is linear with the new points. We developed a KRR-based emulator of PROSAIL to generate 1,000,000 pairs of Sentinel-2 spectra (13 spectral channels) and 7 associated parameters: Total Leaf Area Index (LAI), Leaf angle distribution (LAD), Solar Zenit Angle (SZA), Azimut Angle (PSI), Chlorophyll a+b content $C_{ab}$ [$\mu$g/cm$^2$], equivalent water thickness $C_w$ [g/cm$^2$] and dry matter content, $C_m$ [g/cm$^2$]. See Table~\ref{tab:prosail} for some configuration details of the emulation runs. Some numbers should be given here to truly appreciate the power of emulators: even for the fast PROSAIL, generating the original $5000$ training samples took 20 minutes, training the KRR model/emulator took around 5 minutes, and generating the one million dataset from the trained model (emulator) took few seconds in a standard laptop.


\begin{table}[h!]
\renewcommand{\tabcolsep}{1.2pt}
\caption{Configuration parameters of the simulated data.\label{tab:prosail}}
\begin{center}
\begin{tabular}{|l|l|l|l|}
\hline
\hline
Parameter &	Sampling & Min & Max \\
\hline
\hline
\multicolumn{4}{|l|}{RTM model: Prospect 4} \\
\hline
Leaf Structural Parameter            				&	Fixed                      	& 	1.50	& 1.50 \\
C$_{ab}$, chlorophyll a+b [$\mu$g/cm$^2$]  & 	${\mathcal U}(14,49)$      	& 	0.067	& 79.97 \\
C$_w$, equivalent water thickness [mg/cm$^2$] 			&	${\mathcal U}(10,31)$ 	&	2	& 50 \\
C$_m$, dry matter [mg/cm$^2$]         			& 	${\mathcal U}(5.9,19)$ 	&	1.0 	& 3.0 \\
\hline
\multicolumn{4}{|l|}{RTM model: 4SAIL} \\
\hline
Diffuse/direct light    &	Fixed         &                                      	10 &	10\\
Soil Coefficient       & 	Fixed         &                                      	0 &	0\\
Hot spot               &	Fixed         &                                      	0.01 &	0.01\\
Observer zenit angle     &	Fixed         &                                      	0 &	0\\
LAI, Leaf Area Index    &	${\mathcal U}(1.2,4.3)$ &   0.01 &	6.99\\
LAD, Leaf Angle Distribution  &	${\mathcal U}(28,51)$ &   20.04 &	69.93\\
SZA, Solar Zenit Angle        &	${\mathcal U}(8.5,31)$  &   0.082 &	49.96\\
PSI, Azimut Angle             &	${\mathcal U}(30,100)$ &    0.099 &	179.83\\
\hline
\hline
\end{tabular}
\end{center}
\end{table}

\begin{figure}[t!]
\centerline{\includegraphics[width=16cm]{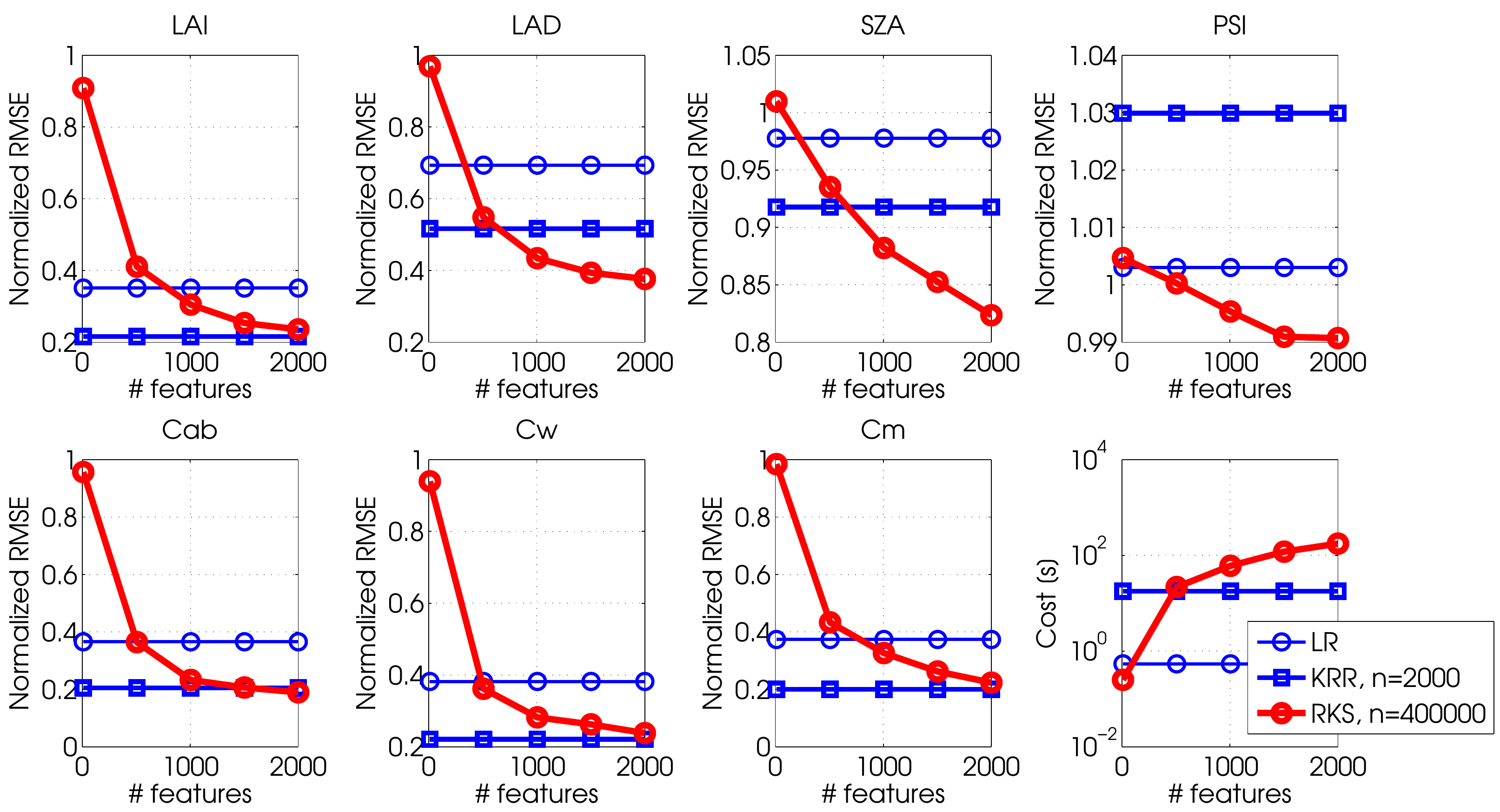}}
\caption{RMSE results in the PROSAIL inversion experiment for the seven parameters and the computational cost (bottom right).}
\label{fig:prosail}
\end{figure}

The developed emulator mimics the complex RTM by learning the input-output nonlinear relations directly from data. By doing so, emulators encode in a set of weights the physical rules governing the vegetation-canopy interactions in PROSAIL. 
The one-million spectra dataset generated was now used for model inversion. Figure~\ref{fig:prosail} shows the obtained results for the inversion of the PROSAIL emulator. We show both the normalized RMSE and the computational cost of a regularized linear regression, KRR and RKS. In all cases we predict the seven parameters with a single multiple-output regression model. In this experiment, we trained KRR with 2,000 samples, and consequently trained RKS for a maximum of $D=2000$ random features for the sake of a fair comparison. RKS employed 400,000 samples for training and cosine basis. Several conclusions can be derived: 1) RKS yields in general competitive performance versus KRR; and 2) RKS largely improves predictions for LAD, SZA, and PSI estimation, while similar in accuracy to KRR for the rest of parameters. 

\subsection{Experiment 3: Cloud detection over landmarks in MSG/SEVIRI}
\begin{figure}[t!]
\centerline{\includegraphics[width=15cm]{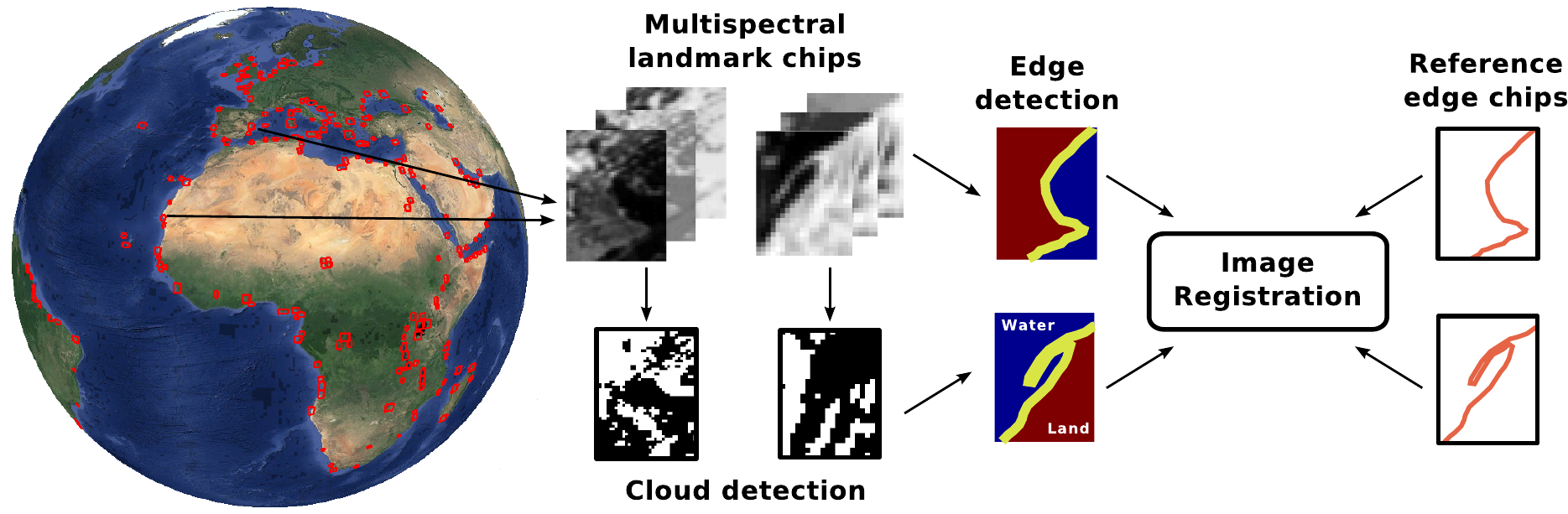}}
\vspace{-0.25cm}
\caption{Landmarks are essential in image registration and geometric quality assessment. Any misclassification of a landmark due to cloud contamination degrades the correlation matching which is a cornerstone for the image navigation and registration (INR) algorithms.}
\label{fig:motivation}
\end{figure}

In the third experiment, we cast the problem of cloud identification over landmarks on Meteosat Second Generation (MSG) data. This satellite mission constitutes a fundamental tool for weather forecasting, providing images of the full Earth disc every 15 minutes. Matching the landmarks accurately is of paramount importance in image navigation and registration (INR) models and geometric quality assessment (GQA) in the Level 1 instrument processing chain. Cloud contamination detection over landmarks is a essential step in the MSG processing chain, as undetected clouds are one of the most significant sources of error in landmark matching (see Fig.~\ref{fig:motivation}). 

The landmark matching application 
requires only a binary detection. 
Furthermore, the cloud detection has to be carried out in real-time to be included in the landmark matching MSG processing chain, which implies efficient and robust detection schemes. Therefore, the proposed classification scheme is designed evaluating the complexity, the scalability, and parallelization of computations.

The dataset provided by EUMETSAT contains MSG/SEVIRI Level 1.5 acquisitions for 200 landmarks of variable size for a whole year, which are mainly located over the coastline, islands, or inland waters. A full Earth disk image is acquired every 15 minutes, which produces 96 images per day and results in 35040 images (or chips) per landmark in 2010. Additionally, Level 2 cloud products were provided for each landmark observation so the  Level 2 cloud mask \citep{Derrien05} is used as the best available `ground truth' to validate the results. Summarizing, in this problem, we have to deal with near 7 million MSG/SEVIRI multispectral images acquired during 2010.

The proposed cloud detection methodology is based on an ensemble of dedicated classifiers. We follow a divide-and-conquer strategy where specific classifiers per landmark and illumination conditions are developed. This strategy allows us to train pixel-based  classifiers with millions of samples at computational affordable times. 
Standard steps in a pattern recognition problem are followed: (1) pre-processing, (2) feature extraction, (3) sample selection, (4) classification, and (5) eventual combination of the individual decisions of a set of trained classifiers in order to obtain the optimal classification ensemble.

First, we perform a conversion from observed calibrated radiance to top of atmosphere (TOA) reflectance and brightness temperature units. Although in statistical retrieval these types of transformations do not dramatically help the learning models, we correct all images to compensate for illumination differences due to diurnal or seasonal cycles. It was necessary because we develop specific classification models for different ranges of sun zenith angle (SZA). 

The next step is feature extraction, where different features are selected as inputs to the classifiers: 7 channels converted to TOA reflectance (R1, R2, R3, R4) and brightness temperature (BT7, BT9, BT10), 3 band ratios, and 6 spatial features (mean and standard deviation of bands R1 and BT9). The informative band ratios are \citep{Derrien05, Hocking11}: a cloud detection ratio, ${\text{R}_{0.8\mu m}}/{\text{R}_{0.6\mu m}}$; a snow index, $({\text{R}_{0.6\mu m}-\text{R}_{1.7\mu m}})/({\text{R}_{0.6\mu m}+\text{R}_{1.7\mu m}})$; and the NDVI, $({\text{R}_{0.8\mu m}-\text{R}_{0.6\mu m}})/({\text{R}_{0.8\mu m}+\text{R}_{0.6\mu m}})$.

Some of the selected classifiers for benchmarking can only ingest a reduced number of training samples due to computational constraints. Hence, sample selection is a critical issue that directly affects the performance of the trained classifiers. We adopt different strategies to alleviate this issue, and also accounting for the land-cover types in each landmark. For each landmark, we select samples that cover all months/dates with a balanced number of cloud-free and cloudy over land and water. For all the analyzed sub-problems, we split the labeled dataset into two disjoint sets: the so-called training and testing sets with different sizes for training (between $5000$ to $10^6$) and $10^6$ pixels for testing.

Finally, in order to simplify the classification task, the different illumination conditions have been independently analyzed splitting the day in four ranges (sub-problems) according to the solar zenith angle (SZA) values: high light conditions (midday), medium light conditions, low light conditions (sunrise/twilight), and night.
Therefore, the final implemented classification scheme considers a pool of 4 classifiers per landmark, each one of them dedicated to different SZA ranges. We selected the SVM and RKS classifiers trained through the standard $v$-fold cross-validation. 
In the case of the RKS, we use Fourier basis functions for all the cloud detection experiments. 
Note that the SVM classifier is very computationally demanding and we restricted the training set to a limited number of samples ($n\leq10^4$), while the proposed RKS classifier allow us to train the models with higher number of samples ($10^6$), becoming an excellent alternative to SVM in large-scale classification problems. 

\begin{figure}[t!]
\centering
\begin{tabular}{c c}
\includegraphics[width=.45\textwidth]{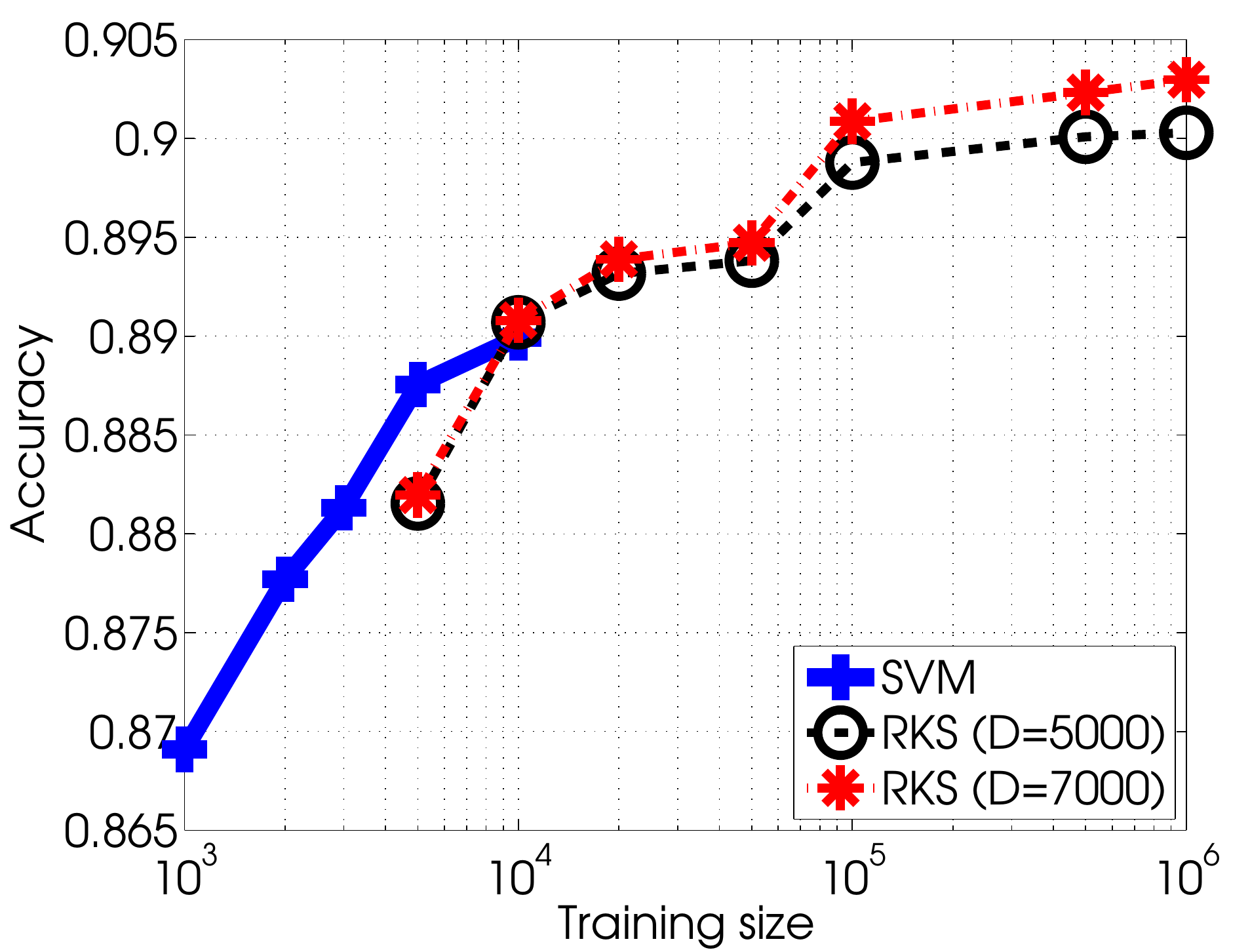} & \includegraphics[width=.45\textwidth]{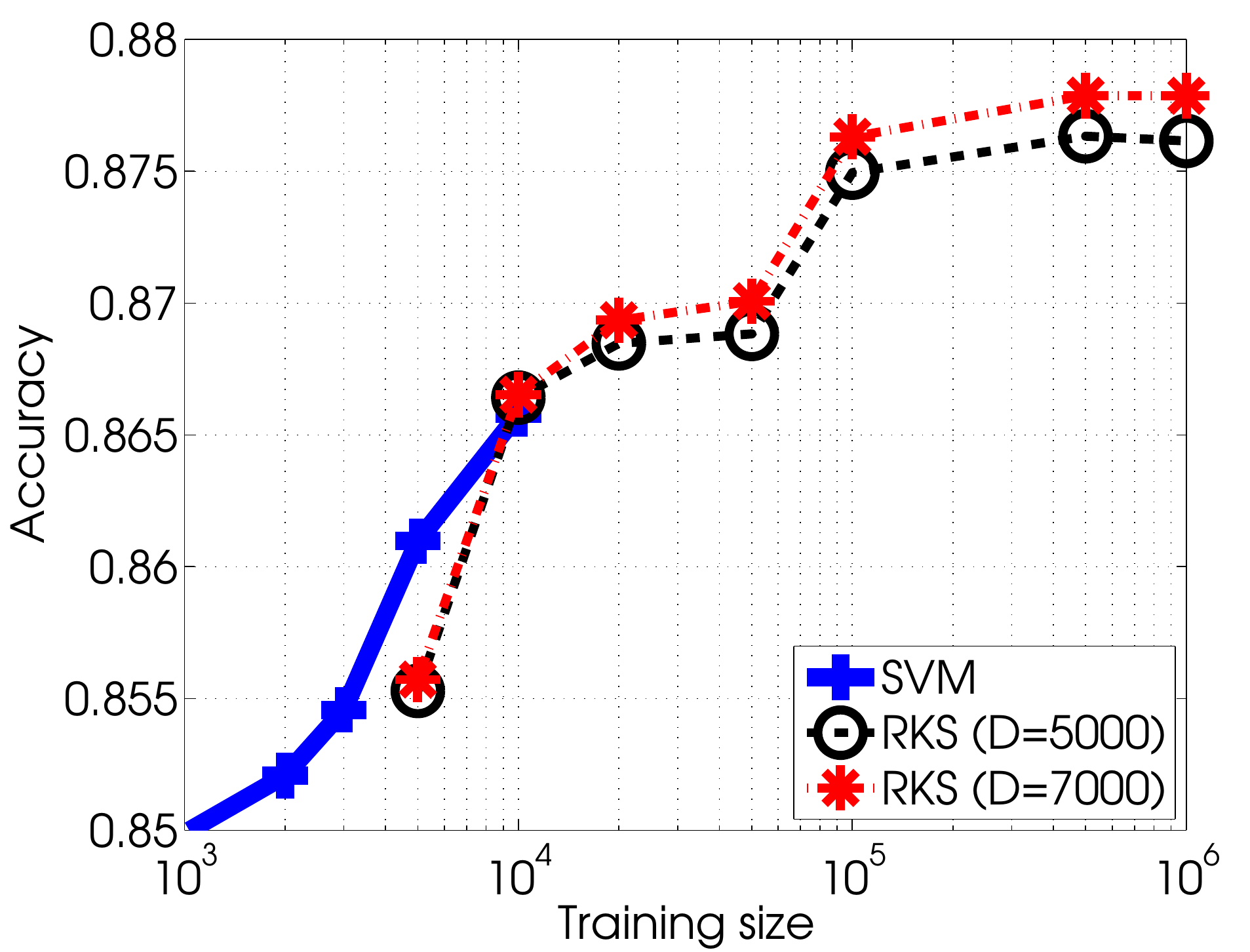} \\
(a) & (b) \\
\includegraphics[width=.45\textwidth]{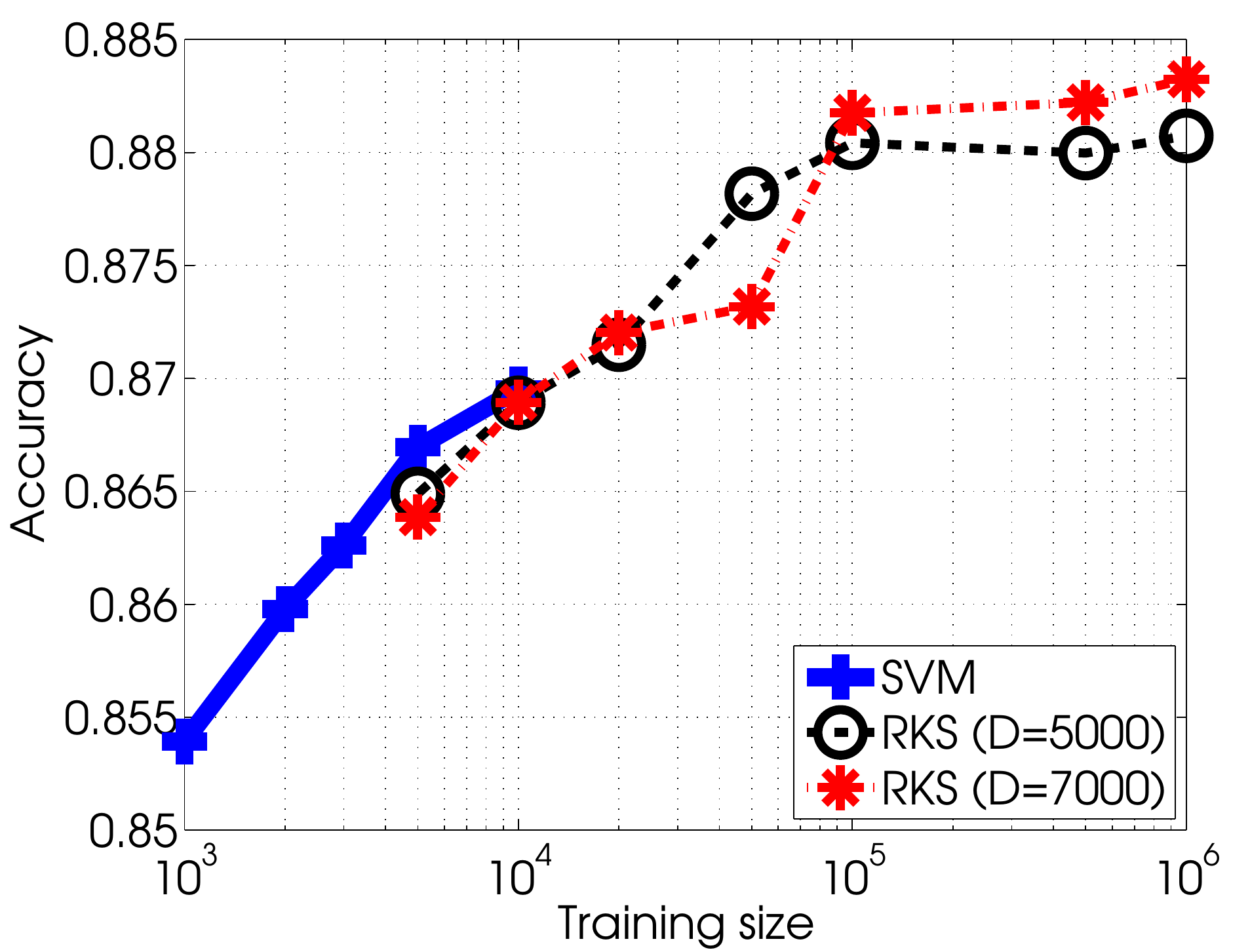} & \includegraphics[width=.45\textwidth]{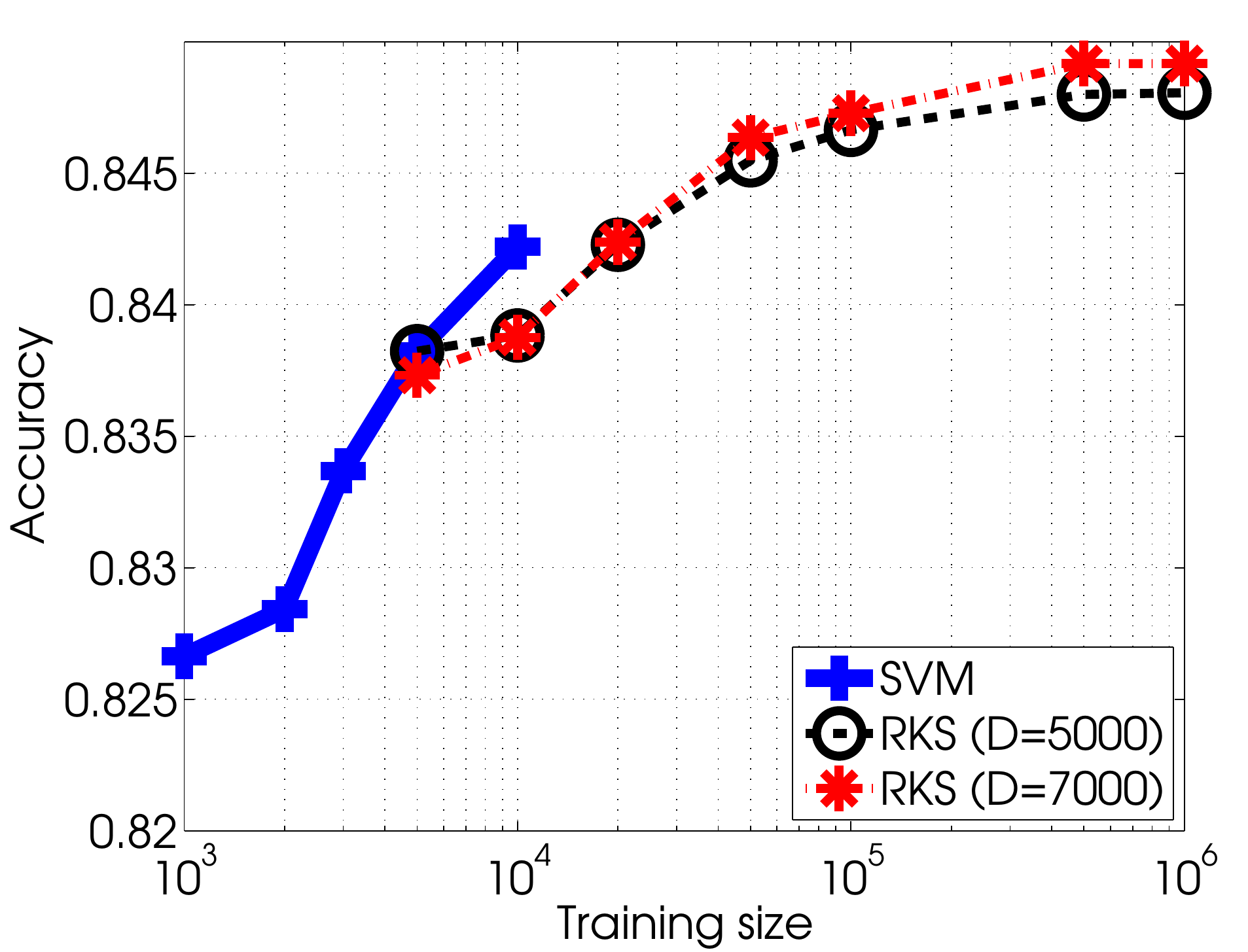}\\
(c) & (d) 
\end{tabular}
\vspace{-0.25cm}
\caption{SVM and RKS classification results (overall accuracy) for the four cloud detection subproblems depending on the light conditions (time-of-day): (a) high light (midday), (b) medium light, (c) low (sunrise/twilight), and (d) night.}
\label{fig:SVM-RKS}
\end{figure}

Classification results for the different time-of-day specific classifiers for the landmark site of Ad Dakhla, Morocco, is shown on Figure \ref{fig:SVM-RKS}. 
We train the SVM classifiers with $[10^3 - 10^4]$, and RKS with $[10^3 - 10^6]$ data points.
We can observe that the best results are reached by RKS classifiers with higher number of training points, despite the SVM outperforms RKS for less than $10^4$ training samples in daytime sub-problems and $20,000$ in the night case.
Hence, the benefits of using RKS are more visible when more information is included ($n>20,000$ training samples). 
It is worth noting that the performance of SVM is good enough and it reach a relatively high overall classification accuracy with a little amount of training data. But the efficiency of RKS dealing with large-scale datasets allows obtaining better results as increases the training set size.  Moreover, the use of more random features, $D$, also provides more flexible solutions increasing the detection accuracy.  However, this increased dimensionality directly affects the CPU time and memory storage, leading to a better results in classification but more computationally demanding. 
The low light conditions (twilight) case and night case are more difficult to solve since one can not rely on visible and near infrared channels. In particular, twilight case has more variability on results due to its complexity. Night is a special case since less features are available to solve the problem (only thermal channels are feed to the classifier), which explain the lower classification accuracy compared to the high and medium light conditions.  

\section{Discussion and conclusions}

This paper explored the use of randomly-generated bases for large scale kernel regression in several remote sensing data processing problems. We focused on the relevant issues of biophysical parameter estimation, model inversion and emulation, and multispectral time series image classification. We exploited the approximation of the kernel function via random sampling from Fourier, wavelets, Walsh and stump functions, and showed results in three relevant problems in Earth observation. First, we tackled a high-dimensional large-scale problem very common in remote sensing: the estimation of atmospheric profiles from large scale hyperspectral infrared sounding IASI radiances. Second, we explored the proposed method for the inversion of the widely used PROSAIL radiative transfer model for which we used 1 million pairs of Sentinel-2 simulations generated via a kernel emulator. Both settings induce multi-input and multi-output problems. The third application dealt with the classification of clouds over landmarks in time series of MSG/Seviri images: we exploited the methodology to train dedicated classifiers for 200 landmarks sites and with different illumination conditions. Results in all problems showed that we can train kernel regression and classification models with several hundreds of thousands of data points, which is not possible in standard kernel optimization strategies, such as support vector machines or kernel ridge regression. 

We noted however that the method has two main shortcomings. First, the memory bottleneck is still present as one has to store the ${\bf Z}$ matrix, which is $n\times D$. This will be addressed in the future through low-rank and block-wise approximations of ${\bf Z}$. And second, other (sparser) bases can be more appropriate. In this work, we used the Walsh basis but results did not improve those of standard Fourier bases. Alternatives to Hadamard expansions, much in line of Fastfood~\citep{Le13}, could eventually improve further the results and efficiency. 

In conclusion, the proposed method produced noticeable gains in accuracy and computational efficiency in all examples. Now it is possible to train sophisticated nonlinear regression methods and classifiers using great many points in a standard laptop. 
The presented framework opens a wide venue to develop more efficient kernel machines in the new Era of big EO data. In this work we focused on the two most relevant problems: retrieval and classification. Nevertheless, it does not escape our notice that the proposed methodology can be applied to other fields of EO data processing (anomaly detection, visualization, clustering, unmixing, feature extraction, etc.) and other data modalities and sensory data (SAR, VHR, etc).

\section*{Acknowledgments}

The research leading to these results has received funding from EUMETSAT under grant agreement EUM/RSP/SOW/14/762293, the European Research Council (ERC) under the ERC-CoG-2014 SEDAL under grant agreement 647423, and the Spanish Ministry of Economy and Competitiveness (MINECO) and FEDER co-funding through the projects TIN2012-38102-C03-01 and TIN2015-64210-R. 

\section*{References}
\bibliographystyle{elsarticle-harv}
\bibliography{random,landmarks,erc}

\end{document}